\documentclass{nle}

\usepackage{amsmath}
\usepackage{graphicx}
\usepackage{natbib}
\ifpdf%
\usepackage{epstopdf}%
\else%
\fi

\usepackage{multirow}
\usepackage{algorithm}
\usepackage[noend]{algpseudocode}
\usepackage{bbold}
\usepackage{ulem}
\usepackage{xcolor}

\usepackage{tikz}
\usetikzlibrary{arrows,positioning,fit,shapes.geometric,backgrounds}

\tikzstyle{gate}=[shape=circle,draw=blue!50,fill=blue!20]
\tikzstyle{var}=[shape=circle, inner sep=0]
\tikzstyle{etc}=[shape=circle,draw=white,fill=white]
\tikzstyle{finalstate}=[shape=circle,draw=blue!100,fill=blue!20]
\tikzstyle{lstm}=[shape=rectangle,draw=orange!50,fill=orange!20,  minimum width=15mm, minimum height=20mm]
\tikzstyle{lightedge}=[<-,dotted]
\tikzstyle{param}=[->,dotted]
\tikzstyle{mainstate}=[state,thick]
\tikzstyle{mainedge}=[<-,thick]

\tikzstyle{bloc}=[shape=rectangle,draw=black!50,align=center]
\tikzstyle{apprentissage}=[shape=ellipse,draw=black!50,align=center]
\tikzstyle{interaction}=[align=center]
\tikzstyle{user}=[shape=rectangle,draw=orange!50,fill=orange!20,align=center]
\tikzstyle{ou}=[shape=diamond,draw=black!50,align=center,scale=0.5]

\tikzset{
    hv/.style={to path={-| (\tikztotarget)}},
    vh/.style={to path={|- (\tikztotarget)}},
}
\usepackage{standalone}
\usepackage{setspace}

\begin{document}
\label{firstpage}

\lefttitle{M. Riou \textit{et al.}}
\righttitle{Findings on On-line Joint Reinforcement Learning of Dialogue Systems modules with real Users}

\papertitle{Article}

\jnlPage{}{}
\jnlDoiYr{}
\doival{}

\title{Findings from Experiments of On-line Joint Reinforcement Learning of Semantic Parser and Dialogue Manager with real Users}

\begin{authgrp}
\author{Matthieu Riou and Bassam Jabaian and St\'ephane Huet and Fabrice Lef\`evre}
\affiliation{LIA, Avignon Universty\\ 339 Chemin des Meinajaries, 84140 Avignon, France\\
            \email{name.surname@univ-avignon.fr}}
\end{authgrp}

\begin{abstract}
Design of dialogue systems has witnessed many advances lately, yet acquiring huge set of data remains an hindrance to their fast development for a new task or language. Besides, training interactive systems with batch data is not satisfactory. On-line learning is pursued in this paper as a convenient way to alleviate these difficulties. After the system modules are initiated, a single process handles data collection, annotation and use in training algorithms. A new challenge is to control the cost of the on-line learning borne by the user. Our work focuses on learning the semantic parsing and dialogue management modules (speech recognition and synthesis offer ready-for-use solutions). In this context we investigate several variants of simultaneous learning which are tested in user trials. In our experiments, with varying merits, they can all achieve good performance with only a few hundreds of training dialogues and overstep a handcrafted system. The analysis of these experiments gives us some insights, discussed in the paper, into the difficulty for the system's trainers to establish a coherent and constant behavioural strategy to enable a fast and good-quality training phase.
\end{abstract}

\maketitle


\section{Introduction}
\label{intro}
While end-to-end deep-learning-based dialogue systems represent a new avenue of research with promising, if not yet definitive, results~\citep{Wen2016b, Wen2017, li2017c}, these models require a huge quantity of data to be trained efficiently. Hitherto, it is not clear how some initial (low cost) knowledge can be leveraged for a warm start of the system development followed by on-line training with users, as described in~\citep{Ferreira2013a, Su2016a}. Although some recent works have proposed end-to-end architectures~\citep{Dhingra2017, Wen2017, Shah2018, Zhao2019}, they still rely on a prior huge data collection before they can reach usable performance. 

So in our work, due to the absence of prior data, the presented system still relies on a classical architecture for goal-directed vocal interaction, with proven capabilities~\citep{Ultes2017}. A cascade of modules deals with the audio information from the user downstream with progressive processing steps. First, words are deciphered from the audio signal (Automatic Speech Recognition, ASR). Then, the utterance meaning is derived (Semantic Parsing, SP) and can be combined with previously collected information, and its grounding status from the dialogue history (belief tracking). On top of it, a policy can make a decision about the following action to perform according to some global criteria (dialogue length, success in reaching the goal, etc.). Dialogue Management (DM) is then followed by operations conveying back the information upstream to users: Natural Language Generation of the dialogue manager action (NLG), then speech synthesis. 

The Hidden Information State (HIS) architecture~\citep{Young2009} offers such a global statistical framework to account for the relations between the data handled by the main modules of the system, allowing for a reinforcement learning of the DM policy. It can be implemented with sample-efficient learning algorithms~\citep{Daubigney2012} and can involve on-line learning through direct interactions with users~\citep{Ferreira2013,  Ferreira2013b, Ferreira2015}. More recently, on-line learning has been generalised to the input/output modules, SP and NLG, with protocols to control the cost of such operations during this human-in-the loop system development (as in~\citep{Ferreira2015e, Ferreira2016a, Riou2017}). The work presented here is a first attempt to combine the on-line learning of SP and DM in a single phase of development. Not only it is expected to help speed-up and simplify the process, it is also likely to benefit from intertwined improvements of the modules.

In dialogue systems, SP extracts a list of semantic concept hypotheses from an input transcription of the user's query. This list is generally expressed as a sequence of Dialogue Acts (DAs) of the form \textit{acttype(slot=value)} and transmitted to the Dialogue Manager (DM) to make the decision on the future action to perform. State-of-the-art SPs are based on probabilistic approaches and trained with various machine learning methods to tag the user input with these semantic concepts~\citep{Lefevre2007, Hahn2011, Deoras2013}. Dealing with supervised machine learning techniques requires a large amount of annotated data which are domain dependent and hardly available.

As a first example of methods to overcome this hindrance, a zero-shot learning algorithm for Semantic Utterance Classification was proposed~\citep{Dauphin2014}. This method tries to find a sentence-wise link between categories and utterances in a semantic space. A neural network can be trained on a large amount of non-annotated and unstructured data to learn this semantic space. Building on this idea, \citet{Ferreira2015} presented a zero-shot learning method for SP (ZSSP) leveragining pre-trained word embeddings~\citep{Mikolov2013}. 

This approach and its extensions~\citep{Bapna2017, Rojas-Barahona2018a} require neither annotated data nor in-context data and have been recently used for different dialogue system modules~\citep{Upadhyay2018, Zhao2018, Bapna2017}. Indeed, the model is bootstrapped with only the ontological description of the target domain and generic word embedding features (learned from large and free general purpose data corpora). But to improve the module further with a light and controlled supervision from the users, an active learning strategy based on an adversarial bandit has also been introduced~\citep{Ferreira2016a}.

In the same line of ideas, made possible by the sample-efficient Reinforcement Learning (RL) algorithm KTD~\citep{Geist2010a}, an active learning scheme has been devised for the training of the DM~\citep{Ferreira2015}. It uses reward shaping~\citep{Ng1999} to take into account local (turn-based) rewards from the user to offer a better control over the learning process and based speed its convergence up.

Stating that solutions exist for active on-line learning of both SP and DM modules, we now consider their simultaneous application to address the issue of the overall system's training. Several challenges must be dealt with in this context. In a modern dialogue system, in contrast to systems handcrafted by experts, many parameters must be learned: contextual mapping of the speakers' words with dialogue acts, as well as the policy of the dialogue manager which aims to choose the best next actions according to a dialogue situation (the history of the dialogue).

Besides, SP and DM intertwine (when one parameter is changed in the semantic parser, it impacts the behaviour of the dialogue manager, and vice versa). So, not only time-saving is foreseen in the joint learning of these two parts of a dialogue system, but also it is expected that good performance will rely on a coherent improvement of both modules.

In this article, a direct application of existing techniques (a bandit algorithm for SP and a RL/Q-learner for DM) is presented and tested; both modules remain separated and the parameters of their on-line training are kept disjoint. Moreover, another possibility with shared parameters in a single Q-learner is introduced and evaluated. This latter presents the pros and cons of an integrated approach: one single unit makes the decision for the two learned modules and can act more coherently, but it has to be fed with larger inputs to base its decision. And in the case of a RL/Q-learner, it can greatly question its performance w.r.t. the training data size, as it will be shown in our experiments.

From a practical point of view, in this work we developed a system intended to be used in a neuroscience experiment. From inside an fMRI scanner, users interact with a robotic platform, vocally powered by the system, which is live-recorded and displayed inside a head-antenna. Users discuss with the system about an image and they tried jointly to elaborate on the message conveyed by the image (see Section~\ref{sec:exp} for further details on this task). The overall architecture of the system in the various configurations tested will be later detailed in Section~\ref{fig:br_rr}. For sake of simplicity, we may note right away that the study is exclusively focused on training SP and DM modules, since well-performing solutions can be used directly off-the-shelf for speech recognition and synthesis.

\section{Related work}
\label{related}
Former works have investigated on-line learning for vocal interaction systems. More generally, this is a general trend in Machine Learning applied to many fields (such as robotics). This idea should not be confused with the general attempt to introduce human-in-the-loop in other NLP tasks, such as Machine Translation where it is used for post-editing~\citep{Koehn2014}, or even already ancient active learning~\citep{Tur2005}.

Specifically, in spoken dialogue systems, early propositions have devised the theoretical foundations of on-line learning, e.g.~\citep{Pietquin2011}. Indeed, training a system with direct interactions required a consequent reduction in need of training data, w.r.t. former solutions based on huge pre-collected datasets or user simulators. Other approaches see on-line learning only as a complementary step on top of traditional training~\citep{Shah2018,Hancock2019} and are more oriented toward user or task adaptation. However, for those willing to fully skip the initial training phase (and its cost) the main difficulty to break down is the slow learning curve of the reinforcement-based algorithms used in the system. Several approaches have helped alleviate this difficulty, such as Gaussian Process modelling~\citep{Gasic2013}, reward shaping alone~\citep{Su2016a}, or combined with Kalman Temporal Differences~\citep{Ferreira2015}, etc. These seminal works have been pursued in several other directions to enlarge the conditions upon which an on-line training could be carried out while ensuring the best ratio between the user's involvement (nature and cost of feedback, manual annotations, etc.) and the level of performance reached, e.g.~\citep{Wang2013, Li2016d, Chen2017c, Chen2017d, Chang2017}.

Joint learning of semantic parsers and dialogue management modules has received a considerable increase in interest since the introduction of end-to-end approaches based on deep neural networks. But even before that, some attempts to jointly train several modules of dialogue systems have been carried out, e.g. joint learning of dialogue management and language generation such as in~\citep{Lemon2011}.

More recently, the development of neural technologies applied to SDS has led to solutions presented as being able to fully train the module pipeline~\citep{Wen2016b}. But it appears that in practice none of the proposed systems could reach good performance while considering all the modules in a single training phase. For instance, even if devised as fully compatible with the model, in \citep{Wen2016b} the semantic belief tracker is not trained with the remainder of the network, and pre-trained modules were used instead. Other works have studied such option~\citep{Yang2016, Padmakumar2017, Rastogi2018, Ye2019}. But even if they are based on sound and efficient propositions they have in common to rely on a huge dataset or a user simulator to reach a good level of performance.

Also the existence of newly proposed pre-trained language generators, such as the Transformers-like BERT~\citep{Devlin2018a} or GPT~\citep{Radford2018a}, represents an interesting option combined with transfer learning to bootstrap dialogue systems~\cite{Wolf2019b}. But such models are not available for all languages (a French version of BERT~\citep{Le2019} was made available after the experiments reported here were done). Besides, how they can be applied to complex dialogue tasks, other than simple chit-chat, is still under investigation.

So the work presented here is a novel proposition that combines joint learning of the two main dialogue system components and direct learning with users. For the first time, it is shown that this approach is doable and can lead to good result in terms of performance. Yet, of course, many degrees of liberty exist in the way it is implemented and taken over by trainers (users involved during the on-line learning phase). From a first set of experiments we endeavour to derive some useful insights on practical realisation of the whole training process.

\subsection{Outline}
The remainder of this paper is organised as follows. After presenting the basis of the on-line learning versions of SP in Section~\ref{sec:zssp} and DM in Section~\ref{sec:dm}, we define the simultaneous on-line learning strategies in Section~\ref{sec:joint}. Section~\ref{sec:exp} provides an experimental study with human evaluations of the proposed approaches, along with an analysis giving us some new insights on the practical implications of on-line learning. We conclude in Section~\ref{sec:concl}.

\section{On-line learning for zero-shot SP}

This section describes how our SP module addresses two issues. Firstly, the zero-shot learning algorithm is able to infer semantic concepts from transcripts using ontological information but no annotated data. Secondly, an adversarial bandit enables an active learning strategy.

\label{sec:zssp}
\subsection{Zero-shot learning spoken language understanding}
\begin{figure}[!t]
\centering
\begin{tabular}{c}
 \includegraphics[width=0.60\textwidth]{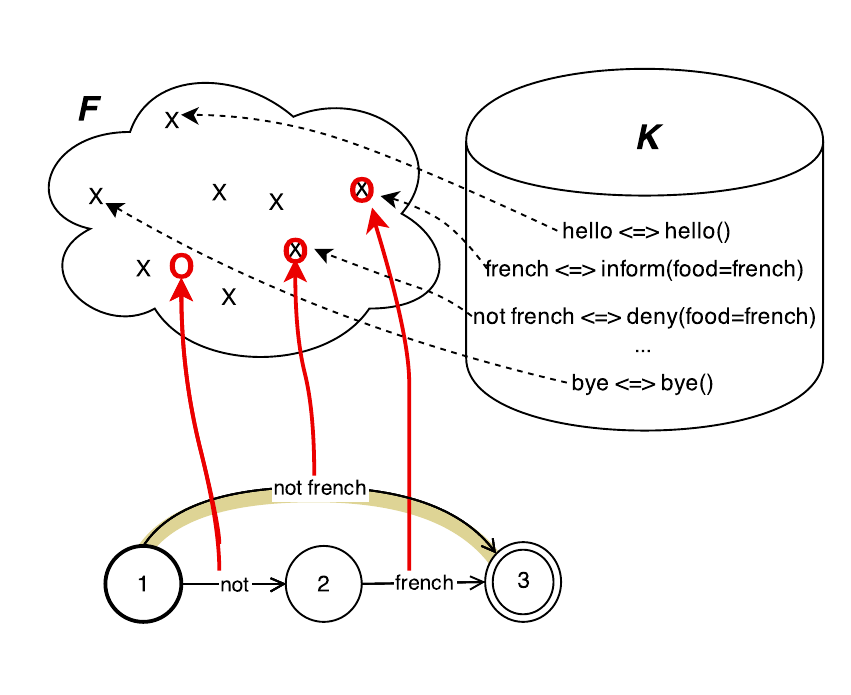}
\end{tabular}
\caption{Basics of the ZSSP Model (from~\citep{Ferreira2016a}): chunks of word-lattice ASR hypotheses are matched with known tuples (surface form, dialogue act, assignment coefficient) in database $K$, through an embedding lexical space $F$}.
\label{fig:zssp}
\end{figure}

The SP model concerned by this study is the ZSSP model presented in~\citep{Ferreira2016a} and illustrated in Figure~\ref{fig:zssp}.  We recalled here only what is necessary to understand its further combination with DM. This model makes use of a semantic knowledge base $K$ and a semantic feature space $F$. $K$ contains collected examples of lexical chunks associated with each targeted Dialogue Act (DA). $K$ is first populated with ontological information (from a back-end database mainly) exclusively, then is completed during the on-line learning process described hereafter. $F$ is a word embedding representation learned with neural network algorithms on large un-annotated open domain data~\citep{Mikolov2013, Bian2014}. 

The ZSSP model builds a scored graph of hypotheses from user utterances. All possible contiguous chunks are considered in the graph and a dot product between the k-most similar vectors and their corresponding assignment coefficients in the $K$ matrix is computed to attribute to each chunk a list of scored semantic hypotheses. 

A best-path decoding is performed in order to find the best semantic tags hypothesis for the considered user utterance.  

\subsection{On-line adaptation based on an adversarial bandit algorithm}
An on-line adaptation strategy (facilitated by the zero-shot approach) is adopted, as presented in~\citep{Ferreira2016a} and briefly recalled here. In this approach, at each dialogue iteration, the system chooses an adaptation action $i_t\in \mathcal{I}$ and updates $K$ according to the user feedback.

The system gain $g(i_t)$, the user effort $\phi(i_t)$ and their combination in the loss function $l(i_t)$ for performing each action are defined and can be estimated during on-line training.   

Three possible actions are considered:
\begin{itemize}
    \item \textbf{Skip}: Skip the adaptation process for this turn ($\phi(\rm{skip}) = 0$).\\
    \item \textbf{AskConfirm}: A yes/no question is presented to the user about the correctness of the selected DAs in the best semantic hypothesis. If the whole sentence is accepted, $\phi(\rm{AskConfirm}) = 1$. Otherwise, $\phi(\rm{AskConfirm})$ is equal to $1+$ the number of DA in the best semantic hypothesis (one yes/no confirmation request per DA).\\
    \item \textbf{AskAnnotation}: the user is asked to re-annotate the whole utterance. \\$\phi(\rm{AskAnnotation}) = 1$ if the sentence is accepted straight away. Otherwise, the user will first inform the system about which chunks she wants to annotate ($+1$ per selected boundary), and then the system will sequentially ask for $acttype$, $slot$ and $value$ if necessary ($+1$ per interim question) for each DA.
\end{itemize}

An adversarial bandit algorithm is used in order to find $i_1, i_2, \ldots, i_t, \ldots$ so that for every $t$, the system minimises the loss $l (i_t)$. The loss function $l(i) \in [0,1]$ is calculated as follows:
$$l(i):=\underbrace{\gamma\, g(i)}_{\rm{system\ improvement}}+\underbrace{ (1-\gamma) \frac{\phi(i)}{\phi_{max}}}_{\rm{user\ effort}}, $$
where $\gamma\in[0,1]$ balances the importance of information improvement and user effort for the system and $\phi_{max}\in\mathbb{N}^*$ is the maximum number of exchanges between the system and the user (in a same turn/round). In this work, $\gamma$ has been set to $0.5$ for example.

\begin{figure}[!t]
\centering
\begin{tabular}{c}
 \includegraphics[width=0.70\textwidth]{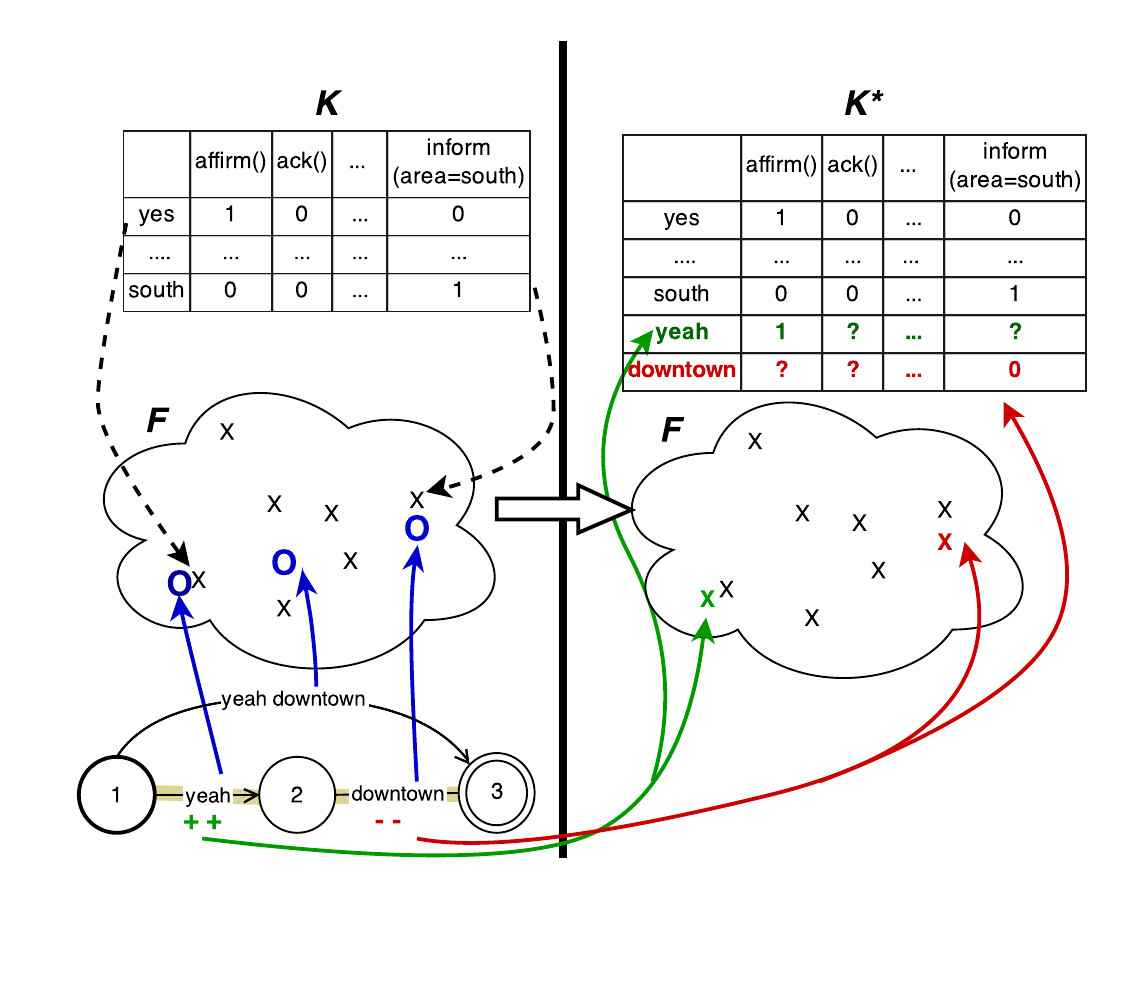}
\end{tabular}
\caption{Adaptation process for the ZSSP Model (from~\citep{Ferreira2016a}): iterative online application of the model to new user data complements $K$ in a controlled way (w.r.t. annotation cost) and improves its coverage and quality. Here users' feedback is only binary, so assignment coefficients are only updated in $K^*$ to (un)validate the couples (surface form, dialog act) tagged by the users }.
\label{fig:zsspadapt}
\end{figure} 

The process is illustrated in Figure~\ref{fig:zsspadapt}. From step at time $t$ on the left, a semantic interpretation is proposed to the user who can annotate it. From the graph we can observe that an AskConfirm has been used in this case: the feedback is expressed in binary terms (correct ``++'' in green, incorrect ``- -'' in red). From these feedbacks the new entries (``yeah'' and ``downtown'') are added to $K$ and their parameters are set accordingly to the feedback values. It should be noted that in that case the information brought by the user is not always complete:
\begin{itemize}
    \item ``++'' means that the proposed act is correct, but it does not mean that it is the only feasible interpretation; in that case, a positive weight associates the entry with the act in $K$, and uncertainty is recorded for the other possible associations. As such, a next use of this line could permit an exploration of other possible associations and gather returns from the user to specify the weight progressively.\\
    \item ``- -'' implies that the act should not be linked to this entry anymore and that the correct association remains to be found. Therefore, a null weight annihilates the correspondence between the new entry and the proposed act. The other parameters are set in a way to explore other possible associations during further appearances of the entry, until some positive confirmation, as in the previous case, could enforce a confirmed pairing.
\end{itemize}

\section{On-line learning for RL dialogue manager}
\label{sec:dm}
The dialogue manager used in this paper relies on the system presented in~\citep{Ferreira2015}. It is based on a POMDP (Partially Observable Markov Decision Process) dialogue management framework, the Hidden Information State (HIS) ~\citep{Young2009}. In a nutshell, the system maintains a distribution over possible dialogue states (the belief state) and uses it to generate an adequate answer. An efficient Reinforcement Learning (RL) algorithm is used to train the system by maximising an expected cumulative discounted reward, with the help of the KTD framework and reward shaping.

\subsection{Policy learning of the dialogue manager within the KTD framework}

At each turn, the dialogue manager generates several possible answers, depending on its belief state. It generates all possible full dialogue acts and matches them to the 11 summary acts 
described in Table~\ref{table:summary_acts} using heuristic rules. Some summary acts can be deemed impossible to realise at some point if no conversion to a full act is possible; for instance, at least one entity must be selected to propose ``Inform''~\citep{Gasic2009}. 
In such a case, a fallback method is implemented which picks the next best possible dialogue act proposed by the policy (knowing that certain actions, e.g. ``repeat", are always possible). Generally speaking, impossibility is due to missing data to convert summary acts into fully-specified acts, and therefore into system's text responses, and ``Ask'' is the only case where the heuristics uses another criterion (repetition) to deem it impossible.

\begin{table}[h!]
  \tbl{\caption{List of the summary acts used by the dialogue manager}
  \label{table:summary_acts}}
 {\begin{minipage}{25pc}
    \begin{tabular}{@{\extracolsep{\fill}}ll}
    \hline
    \textbf{Greet} & Greet user \\
    \textbf{Bye} & End the dialogue \\ \hline
    \textbf{BoldRQ} & Bold query request \\
    \textbf{TentRQ} & Tentative query request \\
    \textbf{Confirm} & Confirm an ungrounded piece of information \\
    \textbf{FindAlt} & Find alternative database entity \\
    \textbf{Split} & Distinguish two hypotheses \\
    \textbf{Repeat} & Repeat \\
    \textbf{Offer} & Offer a database entity \\
    \textbf{Inform} & Give info about current offer \\
    \textbf{QMore} & Query if the user wants more information \\
    \hline
    \end{tabular}
  \end{minipage}}
  {\begin{tabnote}
  \end{tabnote}}
\end{table}

 To learn the policy, i.e. the mapping between situations and actions, an RL approach is used through the KTD learning algorithm~\citep{Geist2010}. This algorithm is derived from a Kalman-based Temporal Differences (KTD) framework, originally devised to track the hidden state of a non-stationary dynamic system through indirect observations of this state. The KTD framework is used in the context of dialogue systems, since it has several advantages and desirable properties for the DM
problem. Indeed, it is sample-efficient, it allows on-policy/off-policy learning through two algorithms (respectively KTD-Q and KTD-SARSA) which
can both perform on-line and offline learning, it provides ways to deal with the ``exploration/exploitation'' dilemma
using uncertainty on value estimates, it allows value tracking, and it supports linear and non-linear parametrisation.
 
 To model the policy of our DM, a linear case was considered. In that respect, the Q-function has a parametric representation $\hat{Q}_\theta = \theta^T \phi(s,a)$ for each state $s$ and action $a$. The
feature vector $\phi(s,a)$ is a set of $n$ basis functions to be designed by the practitioner and $\theta \in \mathbb{R}^n$ the parameter vector
to be learned; the components of the parameter vector $\theta$ are the hidden variables which are modelled as a random vector.
Such parameter vector is considered to evolve following a random walk through an evolution equation: $\theta_t = \theta_{t-1} + v_t$, with $v_t$ a white noise of covariance matrix $P_{v_t}$. In order to train the parameters of the DM system with an off-policy learning, the KTD-Q algorithm is here employed~\citep{Geist2010}.
 
\subsection{Reward function for RL dialogue manager }

At each turn, the policy selects a summary act to answer the user, then feedback is given by the users to score the response, compute a reward, and update the policy. The reward function usually relies on objective criteria such as the success (or not) of the entire dialogue and the number of turns. To take into account local rewards and further speed up the overall training of the system, we also considered reward shaping~\citep{Ng1999}.

As described in~\citep{Ferreira2015}, the reward function is here enhanced with a social reward:
$$ R(s_t,a_t,s_{t+1}) = R_{env}(s_t,a_t,s_{t+1}) + R_{social}(s_t,a_t,s_{t+1}) $$
where $R_{env}$ is an immediate reward function given by the environment (namely a penalty for each turn of the dialogue, aside for the last one which has a reward depending on the success or not of the dialogue), and $R_{social}$ is a potential-based reward shaping function:
$$ R_{social}(s_t,a_t,s_{t+1}) = \lambda\psi_{social}(s_{t+1}) - \psi_{social}(s_t) $$
with $\psi_{social}$ a real-valued function (called the shaping potential function), defined here as a score given by the user at each turn. 
From the user's point of view, there are two different types of feedback. The global feedback is given by the user at the end of the dialogue, depending on whether the entire dialogue is a success or not. The social (or local) feedback $\psi_{social}(s_t)$ is provided at each turn $t$ to score the last response only.
At the end of the dialogue, the policy is updated according to the whole collected feedback.

In this work, $\lambda = 0.95$, the global feedback value is set to 20 in case of success, 0 otherwise. The penalty for each turn is set to -1 and the social feedback $\psi_{social}(s_t) \in\{-1,-0.5,0,0.5,1\}$.

\section{Joint on-line learning}
\label{sec:joint}
\begin{figure}
  \begin{tabular}{cc}
    \tikzstyle{bloc}=[shape=rectangle, rounded corners, draw, align=center, top color=white, bottom color=blue!20]
\tikzstyle{apprentissage}=[shape=ellipse, draw=black!50, align=center, top color=white, bottom color=red!20]
\tikzstyle{interaction}=[align=center]
\tikzstyle{user}=[shape=rectangle, draw=orange!50, fill=orange!20, align=center]
\tikzstyle{ou}=[shape=diamond, draw=black!50, align=center,scale=0.5, top color=white, bottom color=red!20]

\begin{tikzpicture}[node distance=1 cm, thick]

    \node[user] (usr) at (0,0) {User};
    \node[bloc] (asr) at (1,1.5) {ASR};
    \node[bloc] (sp) at (2.5,1.5) {ZSSP};
    \node[apprentissage] (bandit) at (6,1.5) {Bandit};
    \node[ou] (bandit_ou) at (6,0) {OR};
    \node[apprentissage] (rl) at (6,-1.5) {DM \\ {\scriptsize RL policy}};
    \node[bloc] (nlg) at (2.5,-1.5) {NLG};
    \node[bloc] (tts) at (1,-1.5) {TTS};

    \scriptsize
    \draw[-] (bandit) edge [right] node [align=center] {} (bandit_ou);
    \draw[thick,->,>=stealth] (usr) |- (asr) ;
    \draw[thick,->,>=stealth] (tts) -| (usr) ;
    \draw[->,>=stealth]
    (asr) edge [left]  node [align=center] {} (sp)
    (sp) edge [above] node [align=center] {Semantic\\ hypotheses} (bandit)
    (bandit_ou) edge [right] node [align=center] {(Skip)} (rl)
    (rl) edge [] node [align=center] {Action\\ selection} (nlg)
    (nlg) edge [left] node [align=center] {} (tts)
    (bandit_ou) edge [above] node [align=center] {AskAnnotation or AskConfirm} (usr)
    (usr) -- node [right] (a) {Annotation} (sp)
    (a) edge [dashed,below] node [] {Reward$_{SP}$} (bandit)
    (usr) edge [dashed,below] node [] {Reward$_{DM}$} (rl);
\end{tikzpicture}
     &
    \tikzstyle{bloc}=[shape=rectangle, rounded corners, draw, align=center, top color=white, bottom color=blue!20]
\tikzstyle{apprentissage}=[shape=ellipse, draw=black!50, align=center, top color=white, bottom color=red!20]
\tikzstyle{interaction}=[align=center]
\tikzstyle{user}=[shape=rectangle, draw=orange!50, fill=orange!20, align=center]
\tikzstyle{ou}=[shape=diamond, draw=black!50, align=center,scale=0.5, top color=white, bottom color=red!20]

\begin{tikzpicture}[node distance=1 cm, thick]

    \node[user] (usr) at (0,0) {User};
    \node[bloc] (asr) at (1,1.5) {ASR};
    \node[bloc] (sp) at (2.5,1.5) {ZSSP};
    \node[ou] (bandit_ou) at (6,0) {OR};
    \node[apprentissage] (rl) at (6,1.5) {DM \\ {\scriptsize RL policy}};
    \node[bloc] (nlg) at (2.5,-1.5) {NLG};
    \node[bloc] (tts) at (1,-1.5) {TTS};

    \scriptsize
    \draw[-] (rl) edge [right] node [align=center] {} (bandit_ou);
    \draw[thick,->,>=stealth] (usr) |- (asr) ;
    \draw[thick,->,>=stealth] (tts) -| (usr) ;
    \draw[->,>=stealth]
    (asr) edge [left]  node [align=center] {} (sp)
    (sp) edge [above] node [align=center] {Semantic\\ hypotheses} (rl)
    (bandit_ou) |- node [align=center,pos=0.7] {Action selection\\ (not Ask\_)} (nlg)
    (nlg) edge [left] node [align=center] {} (tts)
    (bandit_ou) edge [above] node [align=center] {AskAnnotation or AskConfirm} (usr)
    (usr) -- node [right] (a) {Annotation} (sp)
    (a) edge [dashed,below] node [above] {Reward$_{SP}$} (rl)
    (usr) edge [dashed,below] node [] {Reward$_{DM}$} (rl);
\end{tikzpicture}
  \end{tabular}
  \vspace{1cm}
\caption{Configurations of BR \textit{(left)} and RR \textit{(right)} systems}
\label{fig:br_rr}
\end{figure}
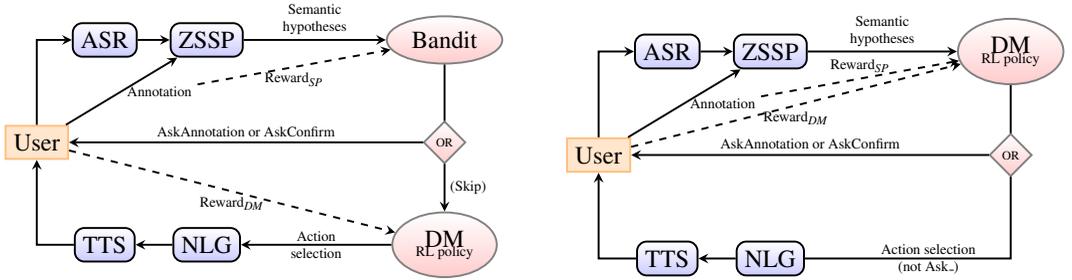

In order to effectively learn the dialogue system on-line, the expert user needs to be able to both improve the SP model and the dialogue manager. Two different joint learning protocols are proposed to achieve it. Both protocols are illustrated in Figure~\ref{fig:br_rr}.

The first one, referred to as \textbf{BR}\footnote{BR stands for \textbf{B}andit-SP and \textbf{R}L-DM learning protocol.} hereafter, directly juxtaposes the bandit to learn the ZSSP and the Q-learner RL approaches to learn the dialogue manager policy. An adversarial bandit algorithm (see Section~\ref{sec:zssp}) is applied for training ZSSP and a Q-learner (see Section~\ref{sec:dm}) is used to learn the DM policy. 
The knowledge base of the ZSSP is updated after each dialogue turn (if the chosen action is not \textbf{Skip}) and the DM policy is updated at the end of each dialogue.

The second protocol, referred to as \textbf{RR}\footnote{RR means that the system is learned with an \textbf{R}L-SP and an \textbf{R}L-DM.} hereafter, directly adds the ZSSP learning actions to the dialogue manager RL policy, and therefore combines the two learning processes into one single policy.

This variant of joint learning merges both policies in a single Q-learner. In that purpose, the DM summary state vector used by the policy was augmented with a ZSSP-related dimension. Let us note that only one dimension was added so as to limit the increase of the state size. At this point, it is handcrafted as no data is available to derive it optimally from raw observations.

This new dimension is evaluated from a set of quality indices of the annotations made by the ZSSP model. A 3-point scale is based upon five distinct features:
\begin{enumerate}
\item \textbf{confidence} $[0,1]$: confidence score of the semantic parser;\\
\item \textbf{fertility} $[0,1]$: ratio of concepts w.r.t. the utterance word length, since ZSSP tends to produce an over-segmentation of the incoming utterances with inserted concepts;\\
\item \textbf{rare} ${0,1}$: presence of rare concepts in the annotation. Rare concepts are ``help", ``repeat", ``restart", ``reqalts", ``reqmore", ``ack" or ``thankyou", and are wrongly annotated in general;\\
\item \textbf{known chunks} $[0,1]$: ratio of annotated chunks available in the semantic knowledge base $K$ among the total number of annotated chunks.\\
\item \textbf{gap} $[0,\infty [$: the difference between the confidence scores of the 1-best and the 2-best annotations. Since those differences are very low ($< 0.01$), the natural logarithm is applied to break out the data in order to have more readable values.
\end{enumerate}

From these features, the ZSSP-related DM state new dimension is computed as:
\begin{enumerate}
\item[0] \textbf{\textit{all clear}}: rare $= 0$ and confidence $\leq 0.499$ and fertility $\leq 0.4$ and known chunks $\geq 0.5$ and gap $\geq -5.5$;\\
\item[1] \textbf{\textit{average condition}}: rare $= 0$ and fertility $\leq 0.5$ and known chunks $\geq 0.15$ and gap $\geq -6.5$ and (confidence $> 0.499$ or fertility $> 0.4$ or known chunks $< 0.5$ or gap $< -5.5$);\\
\item[2] \textbf{\textit{alarming}}: rare $= 1$ or fertility $> 0.5$ or known chunks $< 0.15$ or gap $< -6.5$.
\end{enumerate}

Under the RR protocol, the two ZSSP-annotation actions (``Askconfirm'' and ``Askannotation'', see Section~2) are also included inside the list of summary actions that can be picked up by the dialogue policy. In such a case, the user is presented with the appropriate annotation window in the system's graphical interface and can correct the current annotation. Purely vocal interactions for this process are under study. Yet feasible, it remains a challenging task which could introduce errors of its own, so it seemed more appropriate to evaluate the whole process first with a graphical interface and no input errors. Once done, the turn is updated (i.e. the annotation process has taken the place of the normal user audio response) and the dialogue is pursued. Even though the policy might learn it by itself, we chose to inhibit two ``Ask'' actions in a row (they are tagged as impossible in the next turn), to avoid cycles of annotations. Finally, these two ZSSP-annotation actions have a specific social feedback: instead of $-1$, the feedback $f_i$ uses the loss function $l(i)$ defined in Section~\ref{sec:zssp} and rescaled to obtain a score in $[-1,1]$: 
$$ f_i = (1 - l_i) \times 2 - 1 \enspace .$$

\begin{table*}
\caption{Example of a successful dialogue (translated from French).
}
\label{table:ex_dialogue}
\center
\begin{tabular}{rp{0.60\textwidth}c}
\hline
\textbf{S} & So what about this picture?& \multirow{12}{*}{\vspace{10cm}\includegraphics[width=0.25\textwidth]{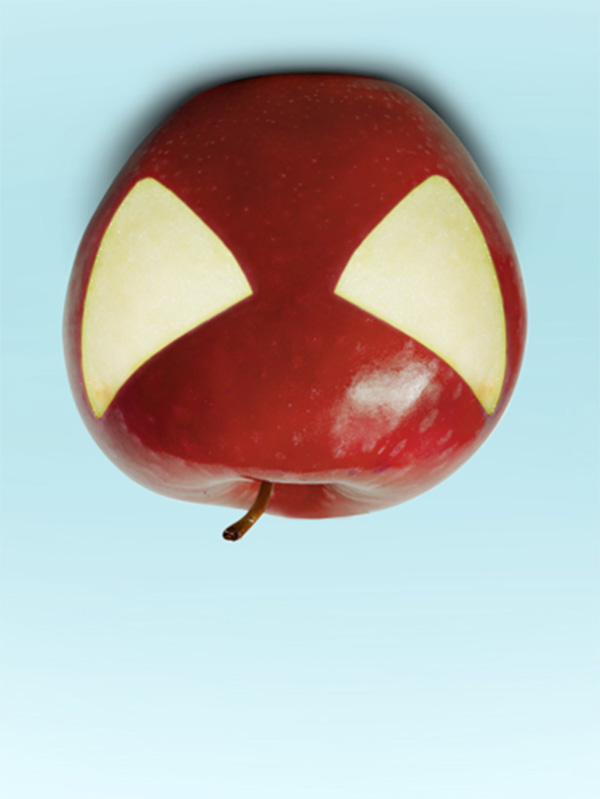}}\\
\textbf{U\_asr} & \textit{I see a red apple.} \\
\textbf{S} & Yes. And what can you say about it? \\
\textbf{U\_asr} & \textit{It's upside-down.}\\
\textbf{S} & It seems to be upside-down. And it's humanised. Due to what? \\
\textbf{U\_asr} & \textit{Because of its eyes.} \\
\textbf{S} & So, it is red and has eyes. Does it make you think of someone? Or does it just look like a basic character? \\
\textbf{U\_asr} & \textit{It looks like spiderman.} \\
\textbf{S} & Indeed, I think this red apple looks like spiderman. It seems it has eyes. It is upside-down.\\
\textbf{U\_asr} & \textit{Can we talk about the message?} \\
\textbf{S} & I think this is related to GMOs.\\
\textbf{U\_asr} & \textit{Ah yes, maybe.} \\
\textbf{S} & And GMOs are genetically modified, like several superheroes.\\
\textbf{U\_asr} & \textit{Yes, they are. Goodbye.} \\
\hline
\end{tabular}
\end{table*}

\section{Experimental study}
\label{sec:exp}
%
\subsection{Task description}
Experiments presented in this paper concern a chit-chat dialogue system framed in a goal-oriented dialogue task~\citep{Riou2019a}\footnote{The datasets generated during the current study are available to the research community upon request to the authors.}. In this context, users discussed with the system about an image (out of a small predefined set of 6), and jointly tried to discover the message conveyed by the image, as described in~\citep{Chaminade2017}. 

The experiments reported hereafter are a step towards a more global objective to develop a system for a neuroscience evaluation of human communication social skills. Inside an fMRI, users will interact with a robotic platform, vocally powered by the presented system, which is live-recorded and displayed inside the head-antenna. These experiments will be performed in French on a new task for which no data are yet available. Therefore, it contrasts with the situation where publicly available corpora can be used~\citep{Williams2013, Casanueva2017}. Likewise, crowd-sourcing is not affordable to realise large-scale data collection, as after a few attempts it seems that no platform (Amazon Mechanical Turk or others) offers enough NLP-skilled workers with a good command of French. 

In order to use a goal-oriented system for such a task, the principle which was followed was to construct, as the system's back-end, a database containing several hundreds of possible combinations of characteristics of the image, each associated with a hypothesis of the conveyed message. During her interaction with the system, it is expected that the user progressively provides elements from the image matching entities in the database. This makes the system select a small subset of possible entities from which it can pick both additional characteristics to inform the user with, and ultimately a pre-defined message to utter as a plausible explanation for the image purpose. This allows the user to speak rather freely about the image for several tens of seconds before arguing briefly about the message. The discussion is expected to last around one minute at most. 

The task-dependent knowledge base used in the experiments is derived from the neuroscience task description of fruit images~\citep{Chaminade2017}, as well as from a generic dialogue information task. The semantics of the domain is represented by 16 different act types, 9 slots and 51 values. The lexical forms used in the natural language generation module to render act types were manually elaborated. A total of around 150 surface forms are initially edited for the basic acts (e.g. \texttt{inform(fruit=\$A) $\rightarrow$ `Right, that's an \$A'}), with on average 2-3 variants per act that the system will pick out randomly. They are automatically combined to produce the possible more complex acts (e.g. \texttt{inform(fruit=\$A, seems=\$B, possesses=\$C) $\rightarrow$ `That \$A with \$C seems rather \$B'}). The final template database amounts to 27k distinct forms. An example of a representative dialogue for the task is given in Table~\ref{table:ex_dialogue}.

Although we are fully aware that the experimental setup is rather distant from more classic tasks of slot-filling and database retrieval, the dialogue system was designed as a goal-oriented system and is eventually close to characteristics expected inside a dialogue platform for more ordinary tasks. Our dialogue system takes into account 13 dialogue acts and 9 slots, which are numbers comparable to what can be found in previous task-oriented systems (e.g. \citet{Budzianowski2018} mention between 6 and 16 act types and, between 3 and 15 slots used in seven domains).

\subsection{User interface}
\begin{figure}[!h]
\centering
\begin{tabular}{c}
 \includegraphics[width=0.80\textwidth]{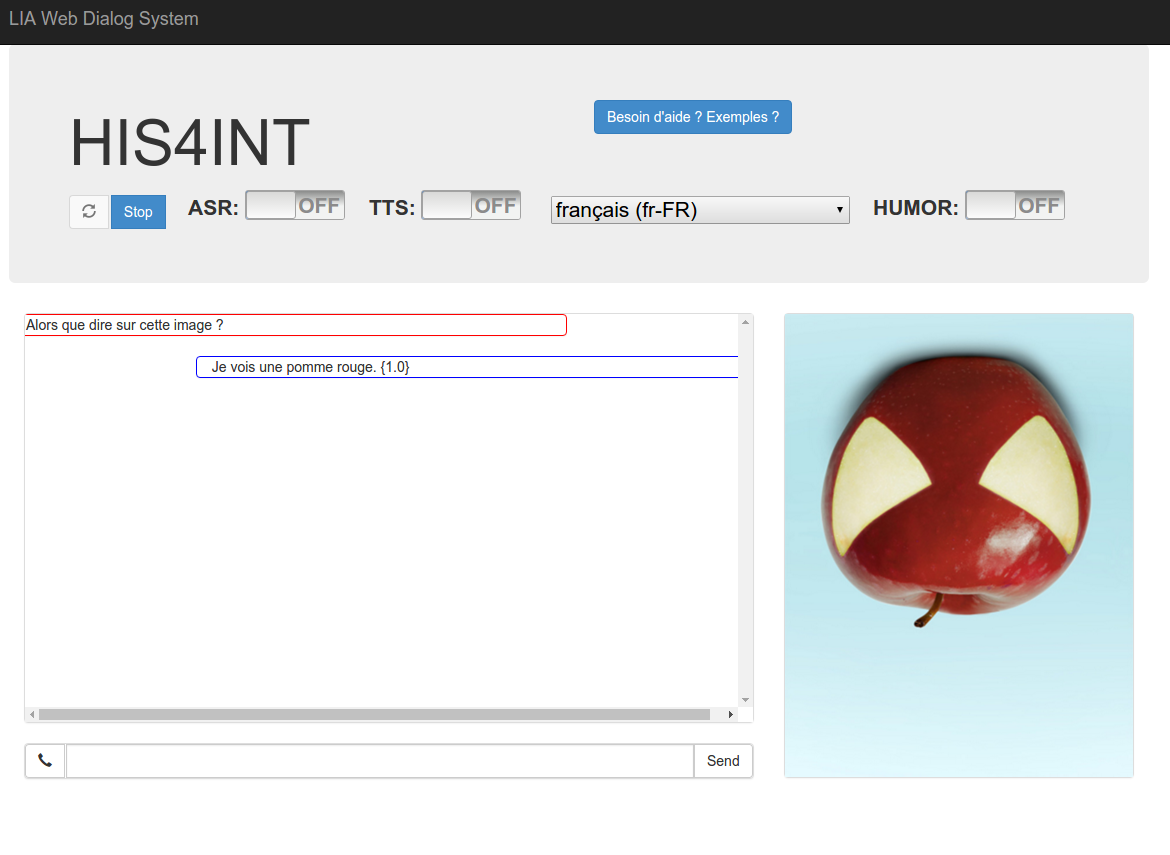}
\end{tabular}
\vspace{1mm}
\caption{User interface: the dialogue window}
\label{fig:interface}
\end{figure}

To handle joint learning of SP and DM, a dedicated user interface has been developed. Basically, it is a web-based interface remotely connected to the dialogue system (see Figure~\ref{fig:interface}). It starts with displaying an information content, then the user can initiate the dialogue and an image appears (randomly selected from the fruit database). 

Before the dialogue starts, the user is prompted with a panel explaining the whole setup. Among the instructions the notion of successful dialogue is detailed:

\noindent ``Once the conversation is over (either you or the system said ``bye"), an evaluation board will appear at the page bottom. A \textbf{successful dialogue} must respect:
\begin{itemize}
    \item at least two image's features have been mentioned (no matter who proposed them);
    \item the system has given its opinion on the message;
    \item no major failure occurred;
    \item and all this in less than a minute!
\end{itemize}
A new dialogue can be started after the board submission."

\begin{figure}[!h]
\centering
\begin{tabular}{c}
 \includegraphics[width=0.80\textwidth]{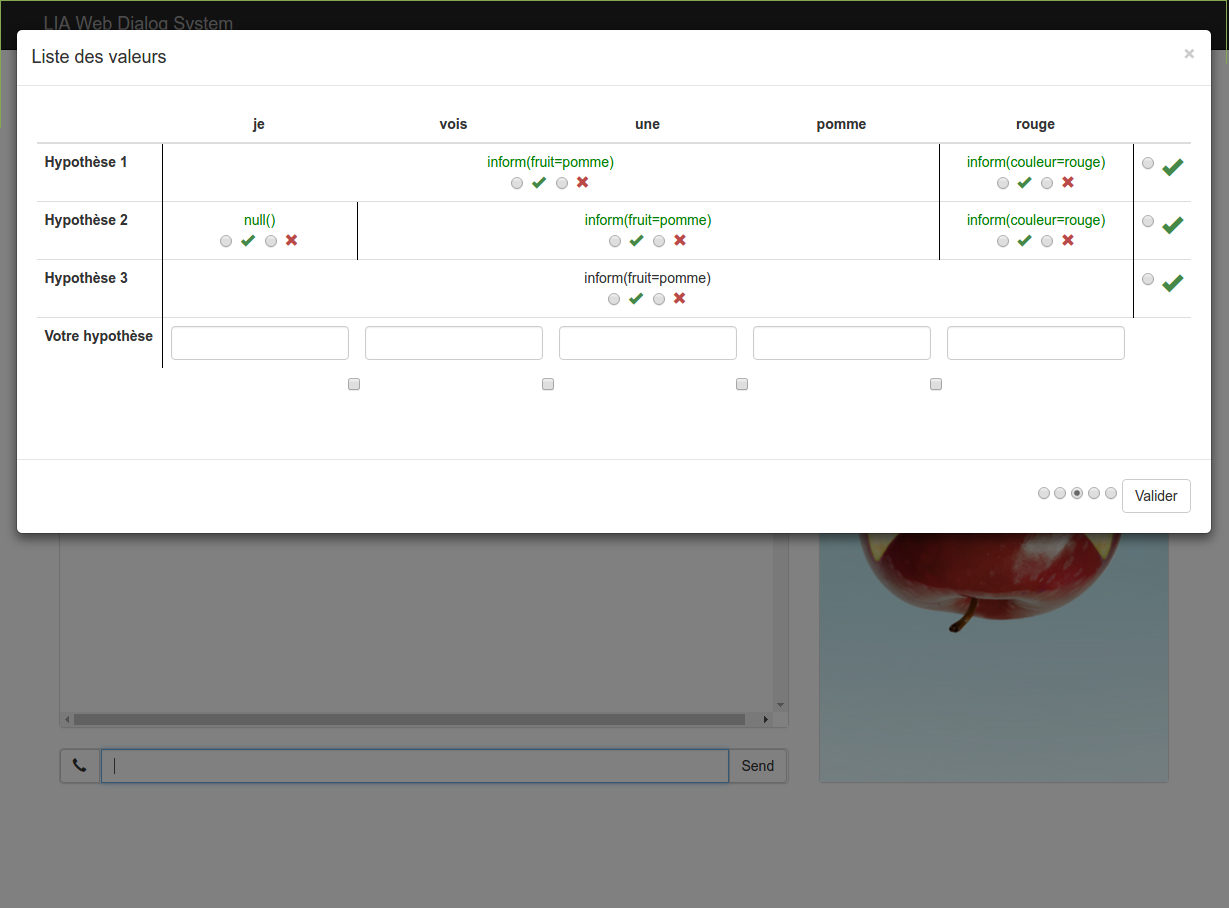}
\end{tabular}
\vspace{1mm}
\caption{User interface: the SP annotation window}
\label{fig:interfacesp}
\end{figure}

\begin{figure}[!h]
\centering
\begin{tabular}{c}
 \includegraphics[width=0.80\textwidth]{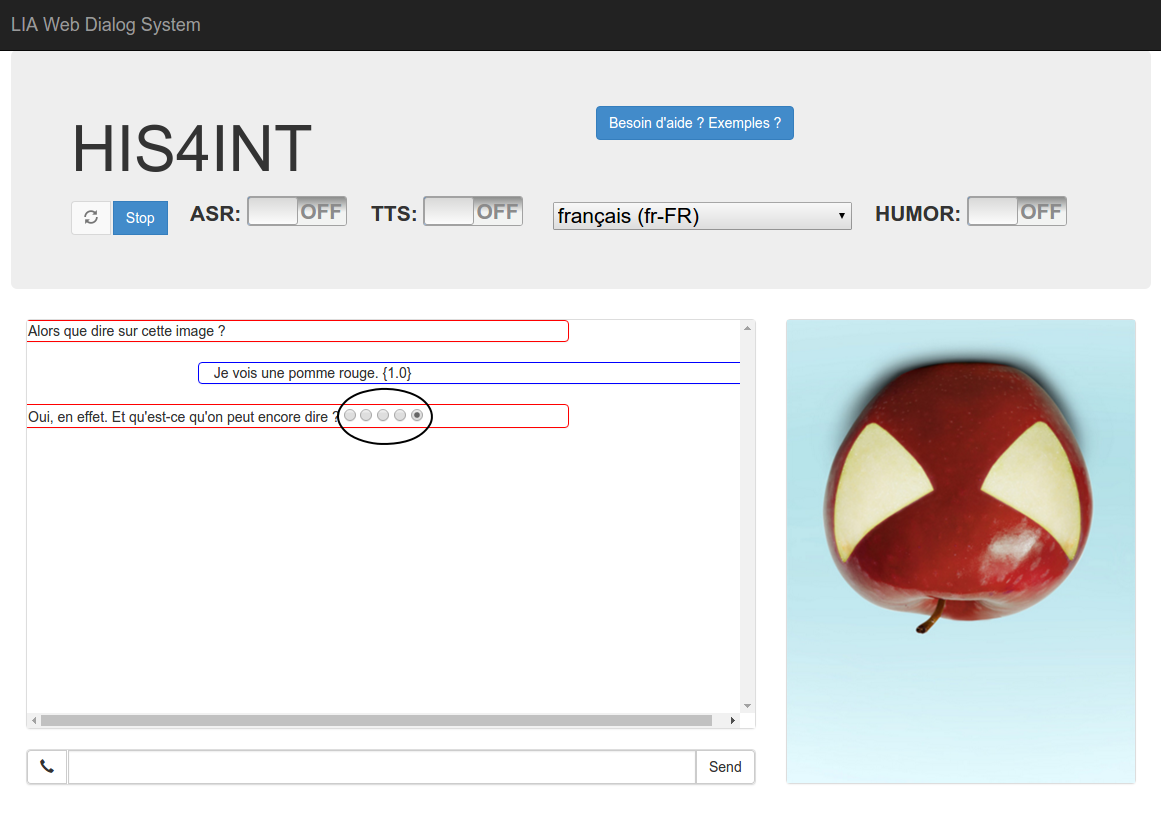}
\end{tabular}
\caption{User interface: turn feedback for social reward of DM}
\label{fig:interfacedm}
\end{figure}

During the dialogue the user is given two possibilities to send feedbacks to the system: 
\begin{itemize}
    \item either through a pop-up window for SP annotations (see Figure~\ref{fig:interfacesp}); in that case the bandit algorithm for ZSSP adaptation, as described in~\ref{sec:zssp}, chooses when a pop-up is triggered and what kind of annotation is asked (only binary feedbacks or full manual annotation);
    \item  or with social rewards for DM reinforcement (see Figure~\ref{fig:interfacedm}); in that case the user decides at each step to use it or not, and when needed to select the value of the reward (in $[-2,2]$).
\end{itemize}

\begin{figure}[!h]
\centering
\begin{tabular}{c}
 \includegraphics[width=0.75\textwidth]{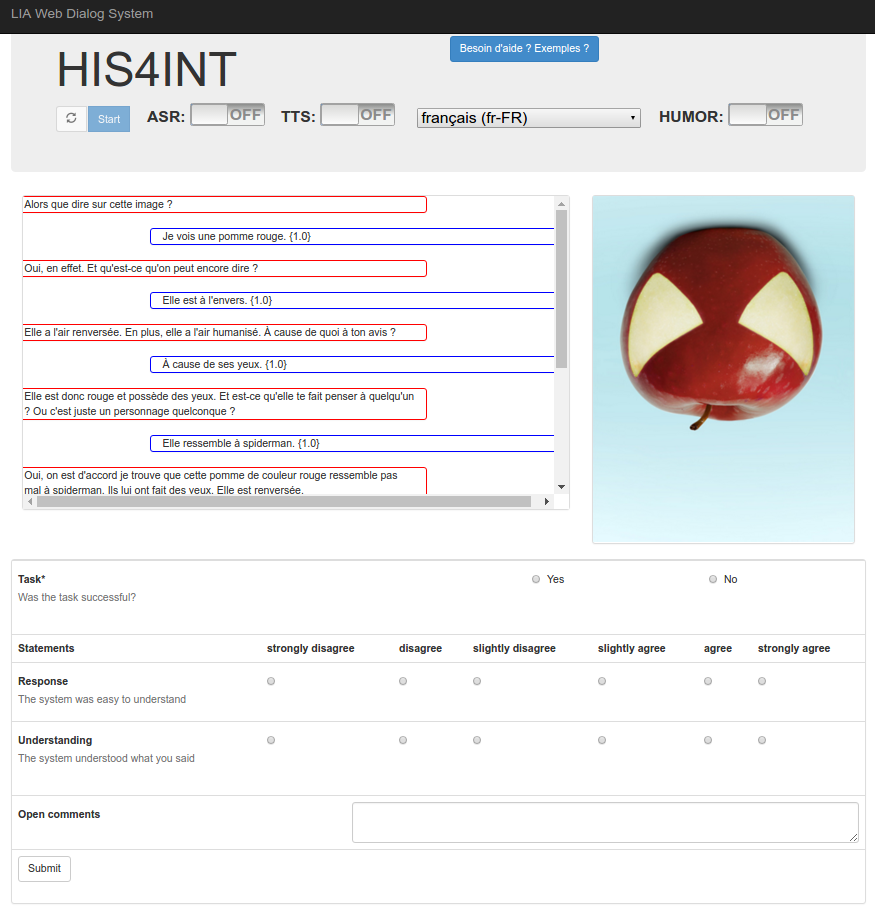}
\end{tabular}
\caption{User interface: final survey}
\label{fig:interfacesurvey}
\end{figure}

After the dialogue has been concluded by either participant, a survey is proposed to collect meta-data  about the success of the interactions and ratings on various dimensions (see Figure~\ref{fig:interfacesurvey} of Annex~1). 

\subsection{Results}
\begin{figure}
  \begin{tabular}{cc}
    \tikzstyle{bloc}=[shape=rectangle, rounded corners, draw, align=center, top color=white, bottom color=blue!20]
\tikzstyle{apprentissage}=[shape=ellipse, draw=black!50, align=center, top color=white, bottom color=red!20]
\tikzstyle{interaction}=[align=center]
\tikzstyle{user}=[shape=rectangle, draw=orange!50, fill=orange!20, align=center]
\tikzstyle{ou}=[shape=diamond, draw=black!50, align=center,scale=0.5, top color=white, bottom color=red!20]

\begin{tikzpicture}[node distance=1 cm, thick]

    \node[user] (usr) at (0,0) {User};
    \node[bloc] (asr) at (1,1.5) {ASR};
    \node[bloc] (sp) at (2.5,1.5) {ZSSP};
    \node[bloc] (rl) at (6,-1.5) {DM \\ {\scriptsize handcrafted policy}};
    \node[bloc] (nlg) at (2.5,-1.5) {NLG};
    \node[bloc] (tts) at (1,-1.5) {TTS};

    \scriptsize
    \draw[thick,->,>=stealth] (usr) |- (asr) ;
    \draw[thick,->,>=stealth] (tts) -| (usr) ;
    \draw[->,>=stealth]
    (asr) edge [left]  node [align=center] {} (sp)
    (sp) -| node [align=center, pos=0.2] {Semantic\\ hypotheses} (rl)
    (rl) edge [] node [align=center] {Action\\ selection} (nlg)
    (nlg) edge [left] node [align=center] {} (tts);
\end{tikzpicture}
     &
    \tikzstyle{bloc}=[shape=rectangle, rounded corners, draw, align=center, top color=white, bottom color=blue!20]
\tikzstyle{apprentissage}=[shape=ellipse, draw=black!50, align=center, top color=white, bottom color=red!20]
\tikzstyle{interaction}=[align=center]
\tikzstyle{user}=[shape=rectangle, draw=orange!50, fill=orange!20, align=center]
\tikzstyle{ou}=[shape=diamond, draw=black!50, align=center,scale=0.5, top color=white, bottom color=red!20]

\begin{tikzpicture}[node distance=1 cm, thick]

    \node[user] (usr) at (0,0) {User};
    \node[bloc] (asr) at (1,1.5) {ASR};
    \node[bloc] (sp) at (2.5,1.5) {ZSSP};
    \node[apprentissage] (bandit) at (6,1.5) {Bandit};
    \node[ou] (bandit_ou) at (6,0) {OR};
    \node[bloc] (rl) at (6,-1.5) {DM \\ {\scriptsize handcrafted policy}};
    \node[bloc] (nlg) at (2.5,-1.5) {NLG};
    \node[bloc] (tts) at (1,-1.5) {TTS};

    \scriptsize
    \draw[-] (bandit) edge [right] node [align=center] {} (bandit_ou);
    \draw[thick,->,>=stealth] (usr) |- (asr) ;
    \draw[thick,->,>=stealth] (tts) -| (usr) ;
    \draw[->,>=stealth]
    (asr) edge [left]  node [align=center] {} (sp)
    (sp) edge [above] node [align=center] {Semantic\\ hypotheses} (bandit)
    (bandit_ou) edge [right] node [align=center] {(Skip)} (rl)
    (rl) edge [] node [align=center] {Action\\ selection} (nlg)
    (nlg) edge [left] node [align=center] {} (tts)
    (bandit_ou) edge [above] node [align=center] {AskAnnotation or AskConfirm} (usr)
    (usr) -- node [right] (a) {Annotation} (sp)
    (a) edge [dashed,below] node [] {Reward$_{SP}$} (bandit);
\end{tikzpicture}
  \end{tabular}
  \vspace{1cm}
\caption{Configurations of ZH \textit{(left)} and BH \textit{(right)} systems}
\label{fig:zh_bh}
\end{figure}
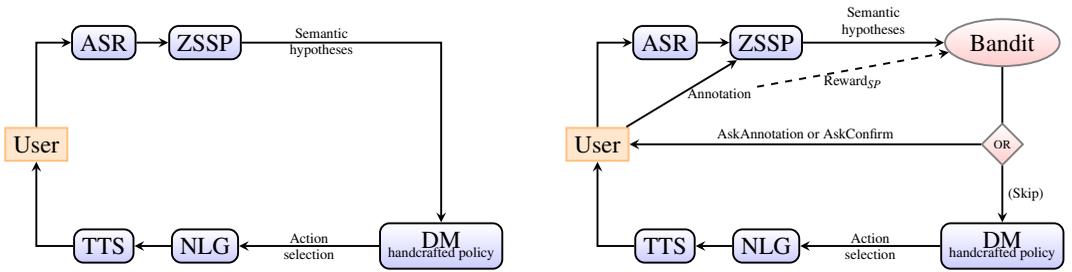

The evaluation of the two joint learning approaches is presented here. Two complementary systems, representing the separate learning classical approach, are proposed in comparison with \textbf{BR} and \textbf{RR} (see Section~\ref{sec:joint}): \textbf{ZH} is a baseline system without on-line learning using the initial ZSSP and a handcrafted dialogue manager policy\footnote{The handcrafted policy used in the system is in line with the description given in sec. 4.4 of \cite{Young2009}.}, whereas the system \textbf{BH} combines the bandit on-line learning for ZSSP and the handcrafted dialogue manager policy. They are both illustrated in Figure~\ref{fig:zh_bh}.

\begin{figure}
  \begin{tabular}{cc}
    \includegraphics[width=0.5\textwidth]{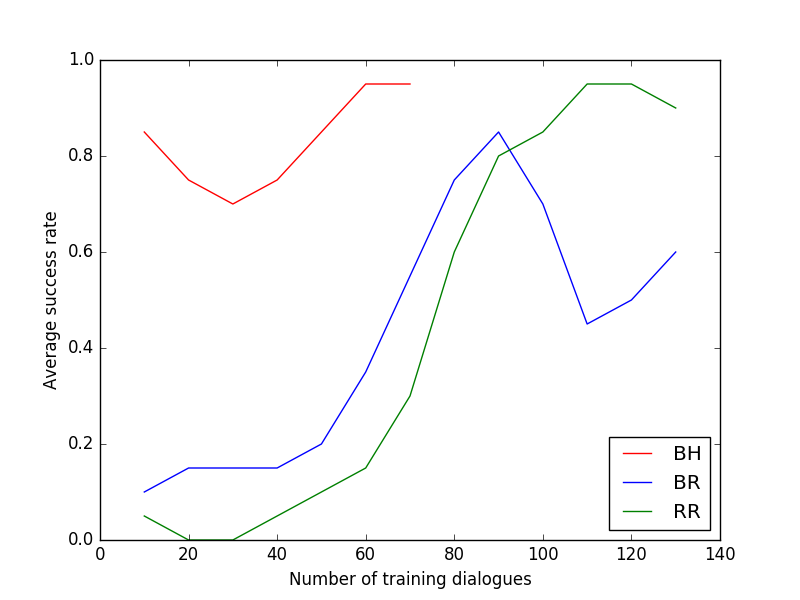}
     &
    \includegraphics[width=0.5\textwidth]{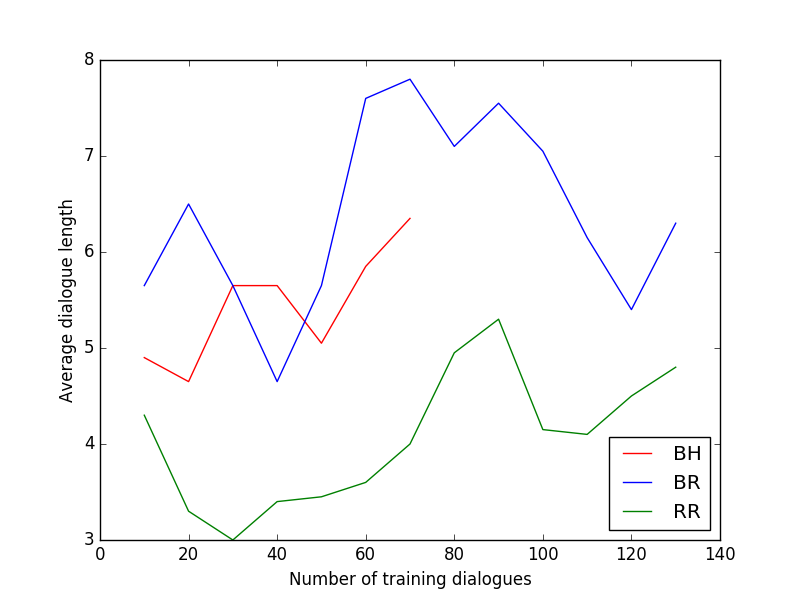}
  \end{tabular}
  \vspace{.3cm}
\caption{\textbf{(a)} Moving average success rates \textbf{(b)} Moving average dialogue lengths}
\label{fig:results}
\end{figure}

For each system, a dialogue system expert user (with a good knowledge of the expected system behaviour and the target task) communicated with the system to train a model. 
Then a group of 13 (mostly) naive users has been recruited to test the models. They were totally unaware of which model they were testing, and models are presented in random order. Due to their availability some have tested several systems, by groups of 12 dialogues per trials. At the end of each session, the users were asked to rate on a scale of 0 (worst) to 5 (best) the understanding and generation qualities of the system. The amount of training dialogues as well as the number of test sets for each configuration are given in Table~\ref{table:evaluation}. Let us note that the number of dialogues (80) to train BH was reduced w.r.t. BR and RR (140), since BH relies on a handcrafted policy, fixed throughout the training phase. Figure~\ref{fig:results} (a) shows that for BH the training average success rate is close to 100\,\% after only 60 dialogues.

\begin{table}[h!]
  \tbl{\caption{Evaluation of the different configurations of joint on-line learning}
  \label{table:evaluation}}
 {\begin{minipage}{25pc}
    \begin{tabular}{@{\extracolsep{\fill}}ccccccccc}
    \textbf{Model} & \textbf{Train} & \textbf{Test} & \textbf{Success} & \textbf{Avg cum.} & \multicolumn{2}{c}{\textbf{Sys. Underst. Rate}} & \multicolumn{2}{c}{\textbf{Sys. Gener. Rate}} \\
    \cline{6-7} \cline{8-9}
    & \textbf{(\#dial)} & \textbf{(\#dial)} & \textbf{(\%)} & \textbf{ Reward} & Mean & SD & Mean & SD \\
    \hline
    ZH & 0   & 142 & 29 & -1.9 & 1.6 & 1.5 & 4.0 & 1.4\\\hdashline
    BH & 80  & 96 & 70 & 7.0 & 3.2 & 1.4 & 4.6 & 0.7\\\hdashline
    BR & 140 & 96 & 89 & 10.9 & 3.3 & 1.6 & 4.6 & 0.7\\\hdashline
    RR & 140 & 96 & 65 & 4.4 & 2.9 & 1.3 & 3.8 & 1.1\\
    \hline
    \end{tabular}
  \end{minipage}}
  {\begin{tabnote}
  \end{tabnote}}
\end{table}

The user trials of the training trials for each protocol are given in Table~\ref{table:evaluation}. For rates computed from the evaluations made by the 13 subjects, mean and standard deviation are provided (last four columns).
The results show that the different configurations of the trained system display acceptable performance. The BR model trained with 140 dialogues shows the best success rate (89\%) and significantly\footnotemark[1] over-performs all other models. Moreover, the ZH model leads to significantly\footnotemark[1] lower success values than all other models. The difference in performance between the ZH and the BH models (+41 points)
shows the impact of the ZSSP adaptation on the overall success of the conversation, along with a better understanding (1.6 for ZH vs. 3.2 for BH). 
Dialogue success annotations have been manually screened after the experiments and no errors w.r.t. to the guidelines given to test users have been noticed.

The average cumulative reward for the test is directly correlated with the success rate and confirms the previous findings. Besides, due to a well-tuned template-based generation system, the system generation quality rate is high ($ \geq 3.8$) for all configurations. The RR protocol offers a success rate smaller than BH and BR (65\% for RR vs. 89\% for BR). RR has been added to the study as it is expected to save some development time, as well as tuning cost in operation. But to achieve this, RR must deal with inputs from both tasks (SP and DM) and process them appropriately to balance the annotation decisions. Clearly the way RR is currently elaborated failed to address this requirement, as will be discussed in more details in Section~\ref{sec:configs}.

\footnotetext[1]{Statistical significance was analysed with a two-tailed Welch's t-test. Results were considered statistically significant with a p-value $ < 0.001$.}

\subsection{Training analysis}

In order to improve our training protocol and to get some insights from our experiments, the training logs were analysed. Only one training is presented for each model in the plots, but in reality several training attempts have been performed by 4 different experts.\footnote{In fact, for each condition's system, two expert users trained two different systems. Due to the difficulty to muster volunteers for the evaluation, only a subset of users was asked to evaluate the two versions of each condition's system. These results are not extensively presented in the paper to avoid interpretation complexity to readers. However, the results are very comparable for the two versions of each system.} After the training process, each expert was asked to describe his training strategy in term of dialogue complexity, utterance lengths and usage of feedbacks turns. Those feedbacks were also analysed in this study. While not statistically significant, these attempts allow us to get a better idea of the properties of on-line training process and the strategy implemented by the expert trainers. The experts were quite free in their training choices. For instance, they did not have restrictions on how they used the additional feedbacks.

\subsubsection{Dialogue complexity: length}
The dialogue complexity is dependent on various factors. Moreover these factors are not independent. Yet we propose to study the impact of dialogue complexity through two main factors: total length in number of turns and amount of exchanged concepts. They are presented separately, in this section and the next, as we posit that the former is better correlated to the evolution of the DM learning when the latter is strongly linked to the SP capacity. Regarding the dialogue manager policy, all experts agree that they preferred to first use simple dialogues to build an efficient dialogue manager policy. Then, when the system started to be usable (regular sequences of 2-3 successful dialogues), more sophisticated dialogues were involved in reaching another step in the system capability. These choices can be seen in the training logs. In Figure~\ref{fig:results} (a) and (b), it can be observed that the success rate and the dialogue length tend to vary a lot during training: 
\begin{itemize}
    \item at the onset of the learning process, dialogues are shorter (with a minimum of 3 turns for RR and around 6 turns for BH and BR) while the success rate is increasing; \\
    \item then, after 40 to 60 dialogues, they become more complex, so that the process drifts towards a decrease of the reward and success rates, and a production of longer dialogues.
\end{itemize} 
These two phases can appear periodically during training. After a phase of complex dialogues, which tends to unsettle the dialogue policy, experts can go back to more simple dialogues for a while.
Let us note that with the BH model, the experts generate sophisticated dialogues faster, since this model does not require to learn a dialogue manager policy in parallel. 

\begin{figure}
  \begin{tabular}{cc}
    \includegraphics[width=0.5\textwidth]{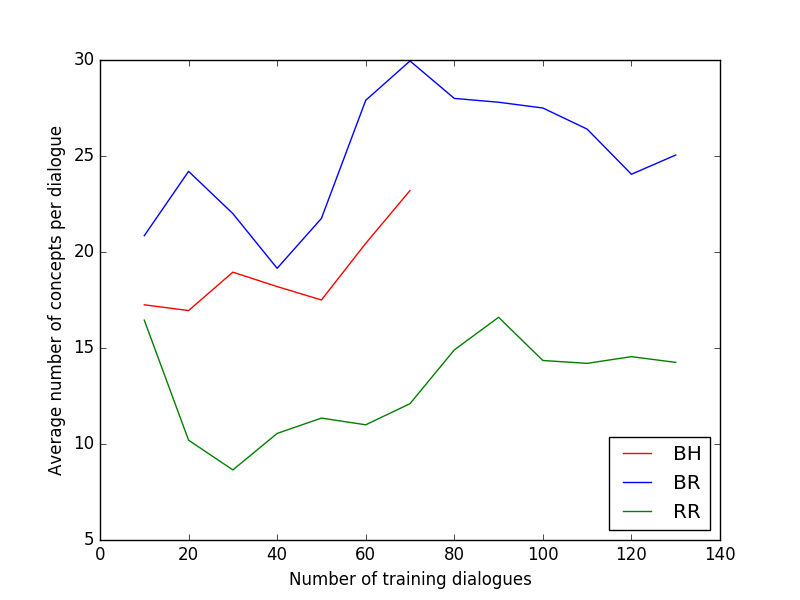}
     &
    \includegraphics[width=0.5\textwidth]{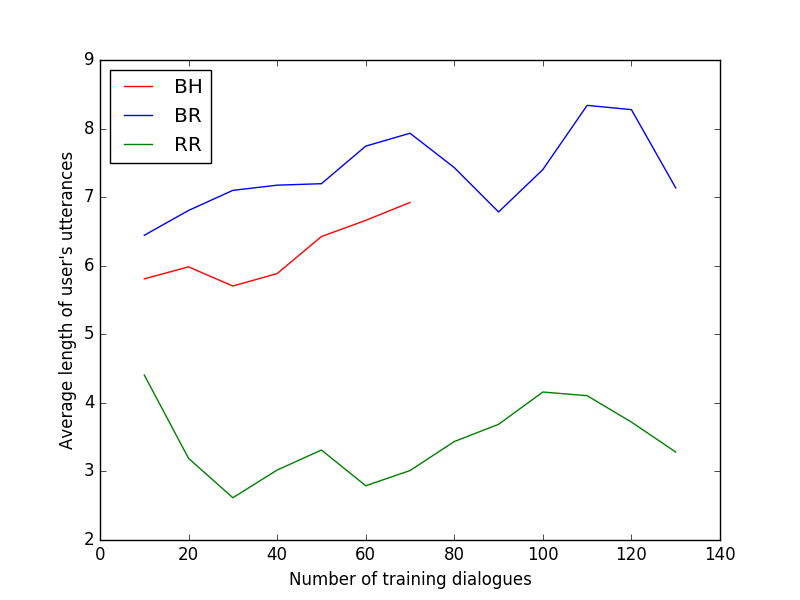} 
  \end{tabular}
  \vspace{.5cm}
\caption{\textbf{(a)} Moving average number of concepts per dialogue \textbf{(b)} Moving average utterance lengths during the training phase for different models}
\label{fig:results2}
\end{figure}

\subsubsection{Dialogue complexity: concepts}
To further analyse ZSSP learning, the total number of concepts exchanged per dialogue during the learning process has been considered. To count the exchanged concepts, each triplet of act, slot and value has been considered as one concept. The results are shown in Figure~\ref{fig:results2} (a). Experts start their training trying their ideal target dialogue. When this one does not succeed, they tend to focus on simpler dialogues with a limited number of concepts. Then, when the system improves, they also diversify the concepts they are using. RR presents fewer concepts per dialogue than BH and BR. This can be explained by the very low triggering level of ZSSP learning actions for this model, therefore the experts cannot efficiently extend the concepts used during a dialogue, and then tend to be conservative in their expressions (and not to introduce too much noise).

This turn complexity can also be observed with the average user's utterance lengths. The results are shown in Figure~\ref{fig:results2} (b). The experts tend to augment their utterance lengths while the system improves, with the exception of RR; in the latter case, the experts tend to have shorter and simpler utterances since ZSSP training is of lower quality.

\begin{figure}
  \begin{tabular}{cc}
    \includegraphics[width=0.5\textwidth]{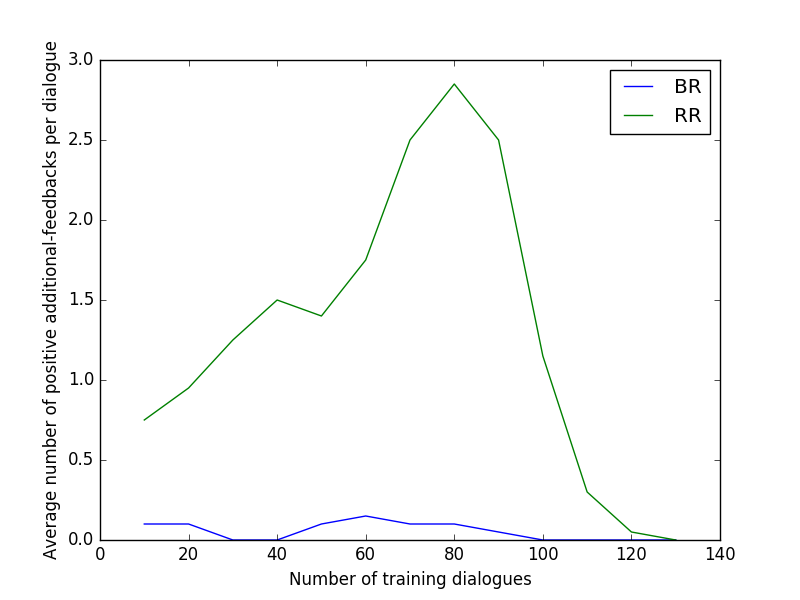}
     &
    \includegraphics[width=0.5\textwidth]{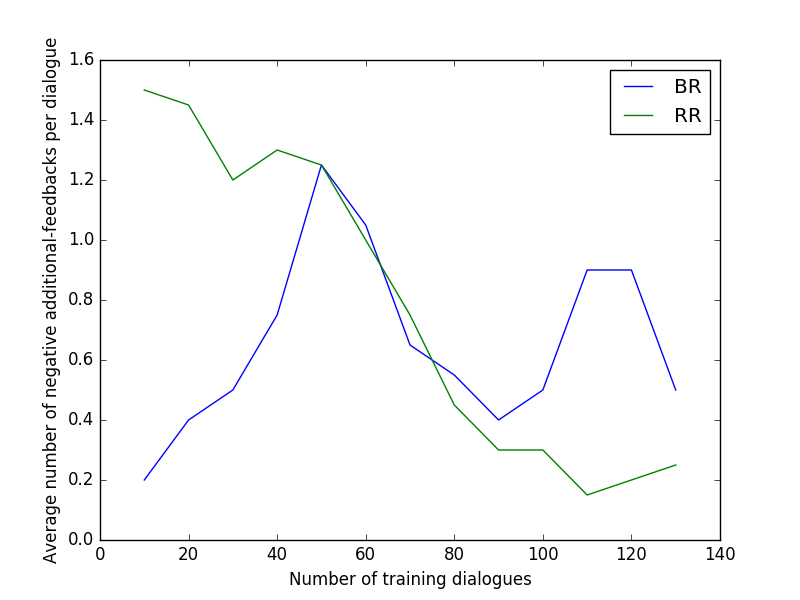}
  \end{tabular}
  \vspace{.5cm}
\caption{\textbf{(a)} Moving average number of additional positive feedbacks per dialogue \textbf{(b)} Moving average number of additional negative feedbacks per dialogue during the training phase for different models}
\label{fig:reward}
\end{figure}

\subsubsection{Dialogue failures}
In order to find possible improvements of the training process, the common causes of dialogue failure were investigated. 48 test dialogues were thoroughly analysed. A large difference of success rates between the models can be explained by the SP errors. An expert was asked to annotate the correct SP outputs for each user input in the test dialogues subset. In addition, an annotation was added when the SP error was due to erroneous speech recognition. Thus, an SP success score was computed as the ratio of the number of correct concepts over the total number of reference concepts plus the number of inserted concepts. ZH presents an SP success rate of 41\% against 70\% for BH, 60\% for BR and 49\% for RR. Those variations can explain the differences of success rates between the different models.

When looking at specific examples of SP errors (see Table~\ref{table:ex_sperrors}), we observe that some of them are common causes of dialogue failures, such as:
\begin{itemize}
    \item a false recognition of a ``goodbye'' act, often due to ASR errors, causing the dialogue to stop prematurely;\\
    \item SP errors creating false beliefs in the dialogue manager, leading to deviate from the image currently being discussed. Those errors are sometimes due to speech errors: about 10\% for ZH, 25\% for BH, 28\% for BR and 27\% for RR. But these differences are to be linked to the SP success rate: a better SP implies fewer errors, so the proportion of SP errors caused by ASR errors increases.
\end{itemize}

\begin{table}[h!]
  \tbl{\caption{Examples of impacting SP errors}
  \label{table:ex_sperrors}}
 {\begin{minipage}{25pc}
    \begin{tabular}{@{\extracolsep{\fill}}rl}
    \hline
    \textbf{ASR (French)}\ \ & il y a des bras des dieux et des jambes \\
    \textbf{ASR (English)} & there are arms gods and legs \\ \hdashline
    & \textit{There is an ASR error due to the phonetic proximity of }dieux \textit{(gods) and }yeux \textit{(eyes).} \\  \hdashline
    \textbf{SP} & inform(looks\_like=superhero,possesses=legs,possesses=arms) \\  \hdashline
    & \textit{The SP misinterpreted }gods\textit{ with the concept of }superhero.\\
    \hline
    \textbf{ASR (French)}\ \ & le fruit a l'air d'avoir baiss\'e les bras \\
    \textbf{ASR (English)} & the fruit seems to have given up \\  \hdashline
    & \textit{The expression} baisser les bras \textit{(to give up) can be translated literally} \\ & \textit{into} lower the arms. \\  \hdashline
    \textbf{SP} & inform(possede=bras), reqalts() \\  \hdashline
    & \textit{Idioms are often difficult to interpret, leading to SP errors. Furthermore,} \\ & \textit{the SP sometimes tends to add acts such as} reqalts() \textit{or} ack() \textit{(especially for} \\ & \textit{ZH and RR).} \\
    \hline
    \end{tabular}
  \end{minipage}}
  {\begin{tabnote}
  \end{tabnote}}
\end{table}

Moreover, negation utterances are still diversely handled by the SP systems. For example ``this is not a lemon'' is not understood in ZH and RR models, but usually is in BH and BR. However, ``this is not a lemon, this is an apple'' remains difficult to address for all models.

\subsubsection{Use of turn feedbacks for DM training}
The additional feedbacks allow the experts to locally reward or penalise specific system responses, in complement to the standard overall reward (task success penalised by length). In~\citep{Ferreira2015}, it has been shown in a simulated environment that negative feedbacks can better guide the dialogue training than positive ones. 

During interviews with the experts after the trainings, they indicate that they seem instinctively more motivated to penalise wrong responses than to reward correct ones,
which is a good thing as a previous study showed in simulations that they were more profitable to the learning process~\citep{Ferreira2015}. However, some experts report they insisted on positive rewards in order to force the system to retain good behaviours. Figure~\ref{fig:reward} shows the uses of both types of feedbacks: negative feedbacks are regularly used for BR and RR (BH policy is handcrafted and does not imply user feedbacks). On the contrary, the use of positive feedbacks is more diverse. BR used almost none while RR used more positive than negative feedbacks. While emitting a lot of positive feedbacks requires more efforts from the expert, it also increases the stability of right choices in the dialogue policy. 

\subsubsection{Comparison of joint learning protocols}
\label{sec:configs}
The main motivation for the RR approach is to rely on a single algorithm. It is expected to save some development time, as well as tuning cost in operation. Yet the two tasks (SP and DM) have distinct representations and a single MDP fails to address both in a same process. 

From the training logs, it can be observed that low performance of RR seems to be related to the very low triggering level of the ZSSP learning actions after the exploration steps during RR w.r.t. the use of the bandit in BH and BR. As a consequence, in RR, the SP learning clearly lagged behind the DM learning, although at onset it is of paramount importance. To remedy this, the policy state space could be modified to take better account of the situations which should lead to ZSSP actions, while preserving its capacities of discrimination for the dialogue actions. Despite our endeavours to add some dedicated indicators in the summary state vectors (see Section 5), it failed to influence enough the value function and policy. But introducing more SP-related features in the Q-learning process is delicate as increasing the size of the state representation vector will impact the sample efficiency of the training algorithm. So in this regard new RL algorithms could be investigated to address the modelisation issue, while maintaining the learning efficiency in the context of very few data.

\section{Conclusion}
\label{sec:concl}
After proposing methods to interactively train both semantic parsing and dialogue management on-line, this paper proposed and evaluated ways to combine them in a joint learning process. Experiments have been carried out in real conditions and are therefore scarce. Yet it has been possible to show that simultaneous learning can be operated, and that after roughly a hundred dialogues the performance of the various configurations tested were generally good enough compared to a handcrafted system. A combination of Bandit for SP and RL/Q-learning for DM oversteps an integrated approach using only RL/Q-learning. While more desirable in terms of development complexity, the latter suffers from merging into a single framework two decision processes based on different ground features.

In any case both configurations offer some insights on the characteristics of the on-line learning process. For instance, in our experiments the experts tend to focus first on simple dialogues to increase success rates, then they try to produce more complex dialogues to both improve the dialogue manager policy and the semantic parser. Our experimental study shows that the use of additional feedbacks at the turn level helps the training. Even if both negative and positive feedbacks can be used, negative ones are commonly preferred to guide the learning process. We also observed that at the end of the training process, most dialogue failures are due to ASR and SP errors. A competitive ASR system remains critical to train an efficient dialogue system.

Based on these results, merging the resulting policies between trials is the next challenge, so as to be able to stack training data coming from different users and save even more time to the system developers.
Another research lead is to allow experts to annotate back their data above the 1-turn current limit. But it would become a post-processing technique, for which the benefit-cost ratio must be evaluated.

\begin{acknowledgements}
This work has been partially supported by grants ANR-16-CONV-0002 (ILCB), ANR-11-LABX-0036 (BLRI).
\end{acknowledgements}

\bibliographystyle{nlelike}
\bibliography{refs}

\newpage
\section*{Annex 1: User interface}

\end{document}